\documentclass[twocolumn]{svjour3}          
\smartqed  

\usepackage{hyperref}
\usepackage{url}
\usepackage{amsmath} 
\usepackage{verbatim}
\usepackage[pdf]{pstricks}
\usepackage{tikz}
\usepackage{breqn}
\usepackage{booktabs} 
\usepackage[utf8]{inputenc}
\usepackage{algpseudocode}
\usepackage{algorithm}
\usepackage{pgfplots}
\usepackage{graphicx}
\usepackage{caption}
\usepackage{multirow}
\usepgfplotslibrary{groupplots}
\usetikzlibrary{fit,positioning}
\usetikzlibrary{patterns}
\usetikzlibrary{bayesnet}
\newcommand{\vect}[1]{\boldsymbol{#1}}
\newcommand{\matr}[1]{\boldsymbol{#1}}

\DeclareMathOperator*{\argmax}{arg\,max}

\newcommand{\variable}[1]{\mathit{{#1}}}

\DeclareMathOperator{\stereo}{stereo}
\DeclareMathOperator{\project}{project}
\DeclareMathOperator{\dist}{dist}
\DeclareMathOperator{\queue}{Queue}
\DeclareMathOperator{\pop}{pop}
\DeclareMathOperator{\push}{push}

\newlength\figureheight
\newlength\figurewidth
\newlength\axislabelsep

\newcommand{\sysname}[0]{$\mathrm{RT}^2$}

\begin{document}

\title{Reliable Real Time Ball Tracking for Robot Table Tennis}


\author{Sebastian Gomez-Gonzalez \and
        Yassine Nemmour \and \\
        Bernhard Schölkopf \and
        Jan Peters
}


\institute{Sebastian Gomez-Gonzalez \at
              Max-Planck Institute for Intelligent Systems \\
              Max Planck Ring 4, Tübingen, Germany \\
              \email{sgomez@tue.mpg.de}           
           \and
              Yassine Nemmour \at
              Max-Planck Institute for Intelligent Systems \\
              \email{ynemmour@tuebingen.mpg.de}
           \and
              Bernhard Schölkopf \at
              Max-Planck Institute for Intelligent Systems \\
              \email{bs@tuebingen.mpg.de}
          \and
           Jan Peters \at
              Technische Universität Darmstadt \\
              Hochschulstrasse 10, Darmstadt, Germany \\
              \email{mail@jan-peters.net}
}


\maketitle

\begin{abstract}
  Robot table tennis systems require a vision system that can track the ball position
  with low latency and high sampling rate. Altering the ball to simplify the tracking
  using for instance infrared coating changes the physics of the ball trajectory.
  As a result, table tennis systems use custom tracking systems to track the ball
  based on heuristic algorithms respecting the real time constrains applied to RGB 
  images captured with a set of cameras. 
  However, these heuristic algorithms often report erroneous ball positions,
  and the table tennis policies typically need to incorporate additional heuristics
  to detect and possibly correct outliers.
  In this paper, we propose a vision system for object detection and tracking that
  focus on reliability while providing real time performance. Our assumption
  is that by using multiple cameras, we can find and discard the errors obtained 
  in the object detection phase by checking for consistency with the positions 
  reported by other cameras. We provide an open source implementation of the proposed
  tracking system to simplify future research in robot table tennis or related tracking
  applications with strong real time requirements.
  We evaluate the proposed system thoroughly in simulation and in the real system,
  outperforming previous work. Furthermore, we show
  that the accuracy and robustness of the proposed system increases as more cameras
  are added. Finally, we evaluate the table tennis playing performance of an existing
  method in the real robot using the proposed vision system. We measure a slight
  increase in performance compared to a previous vision system even after removing
  all the heuristics previously present to filter out erroneous ball observations.
\end{abstract}

\section{Introduction}
\label{sec:intro}
%
%
%
%

\begin{figure}[b]
  \centering
  \includegraphics[width=8.2cm]{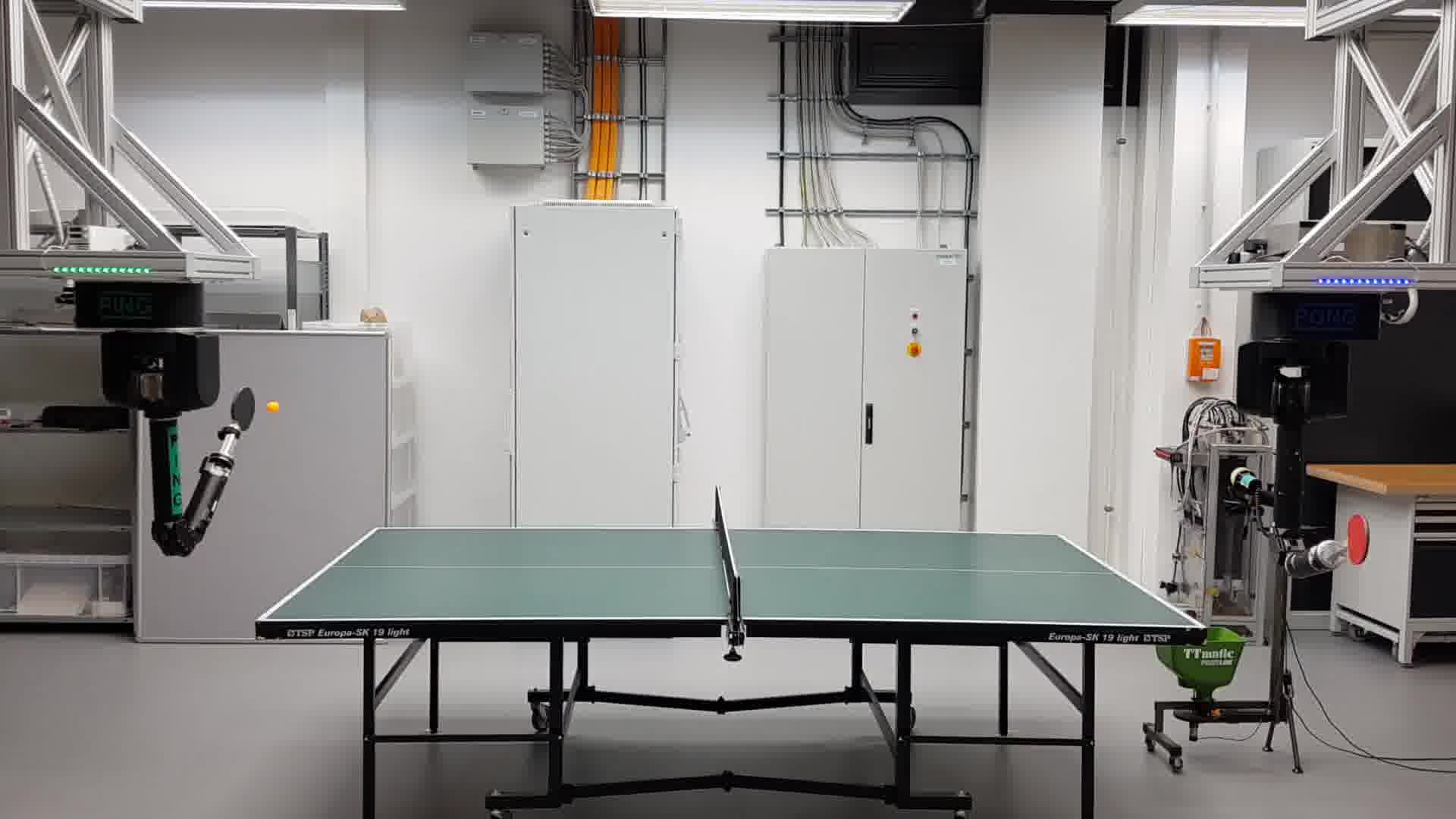}
  \caption{
    Robot table tennis setup used to evaluate the proposed methods. We use four
    cameras attached to the ceiling to track the position of the ball. The robots
    used are two Barrett WAM robot arms capable of high speed motion, 
    with seven degrees of freedom like a human arm.
  }
  \label{fig:robot}
\end{figure}

Game playing has been a popular technique to compare the performance of different 
artificial intelligence methods between themselves and against humans. 
Examples include board games like Chess~\cite{campbell2002deep} and 
Go~\cite{alphago} as well as sports like robot-soccer~\cite{kitano1997robocup}. 
Table tennis has been used regularly as a robot task to evaluate the performance
of ad-hoc techniques~\cite{mulling2011biomimetic}, imitation learning~\cite{gomez2016using},
reinforcement learning~\cite{mulling2013learning} and other techniques in a complex 
real time environment.

In order to play table tennis, a robotic systems needs reliable information about the ball
trajectory with low latency and high sampling frequency. Commercial tracking system like VICON 
can provide reliable 3D positions with high sampling frequencies, but it requires
attaching IR reflective markers to the objects to track. Table tennis balls are very light,
and it is not possible to attach a IR marker or even coat the ball surface with IR reflective
paint without changing the physics of the ball trajectory. For this reason, robot table tennis
approaches typically use software based solutions that take images from a set of video cameras
and estimate the 3D position of the ball. 

Tracking systems for table tennis balls use fast heuristics to detect the ball respecting
the real time constrains required by table tennis systems. These heuristics typically look
for round objects and use color information of table tennis balls. Although these heuristics
work well most of the time, assuming that the reported ball positions are always correct before
the 3D triangulation will result in a number of outliers that increases as more cameras are 
used in the tracking system.

As a result, robot table tennis systems need to incorporate outlier 
detection~\cite{huang2015learning} techniques on the reported 3D positions using for example physical 
models of the ball trajectory~\cite{gomez2016using}. This is unfortunate, 
since it results in effort duplication and reduces the interest of the machine learning
community to work on real robot table tennis platforms.

In this paper, we propose a simple and efficient framework for object tracking. The proposed
framework is tested on a robot table tennis setup and compared with previous work~\cite{rtblob}. 
Unlike previous work, we focus on the reliability of the system without the use of any 
strong assumptions about the object shape or the physics of the flying ball. To evaluate the
performance of the algorithm in setups with different amount of cameras, we use a simulation
environment. We show that adding more cameras helps to increase the robustness and the accuracy
of the proposed system.

In the real system, we evaluate the error distribution of the proposed system and compare it
with previous work~\cite{rtblob}. We show that the proposed framework is clearly superior in
accuracy and robustness to outliers. Finally, we evaluate the system by using a robot table
tennis policy~\cite{gomez2018adaptation} that was designed to be used with the 
RTBlob vision system~\cite{rtblob}. We remove all the heuristics to detect and remove outliers
from the policy implemented~\cite{gomez2018adaptation} and still obtain a slight improvement 
of performance compared using the proposed vision system. Figure~\ref{fig:robot},
shows the real robot setup used on the experiments, executing the policy proposed 
in~\cite{gomez2018adaptation} with the vision system proposed on this paper.

Although we focus on robot table tennis due to its particular real time requirements,
we use machine learning techniques for the object detection part that can be trained to
track different kind of objects. A user only needs to label a few images by placing a bounding box around
the object of interest and train the system with the labeled images.

\subsection*{Contributions}

We provide a open source implementation~\cite{ball_track_impl} of a simple table tennis ball 
tracking system that focuses on reliability and real time performance. The implementation can
be used to track different objects simply by retraining the model. The provided open source 
implementation will enable researchers working on robot table tennis or related real time 
object tracking applications to focus their efforts into better strategies or models, instead
of devising strategies to determine which observations can be trusted and which can not.

We evaluate the proposed system in simulation and in a real robot table tennis platform. In simulation,
we show that increasing the number of cameras results in higher reliability. On the real system,
we evaluate an existing robot table tennis strategy using the proposed vision system with four cameras
attached to the ceiling. The heuristics used to discard outliers on the ball observations where removed,
while obtaining a slightly increase on playing performance. In addition, we provide latency times for
the different experiments to show the proposed system can deliver real time performance even with a
large number of cameras.

\subsection*{Related Work}
\label{sec:intro:ttvision}

Ball tracking systems take an important role in almost all popular ball based sports to
aid coaches, referees and sport commentators. Examples include soccer~\cite{seo1997ball,tong2004effective}, 
basketball~\cite{chen2009physics}, tennis~\cite{pingali2000ball}, etc. 
There are multiple systems designed for tracking table tennis balls, some of which include
real time considerations or were designed for robot table tennis. Table tennis is a fast game, 
and that makes it a hard robot problem to tackle. A smashed ball takes about 0.1 seconds
to reach the other end of the table, and even at beginner it takes about 1 second to the
ball to reach the opponent. Considering that robot arms like the Barrett WAM are much slower
than a human arm, the amount of time available to make a decision of how and where to move
before it is too late to reach the ball is low even to play at a beginner level. As a result,
a vision system for robot table tennis needs to provide a high sampling rate with a low latency
to provide as much information as possible as early as possible.

RTblob~\cite{rtblob} was one of the first vision systems used for robot table tennis applications.
It uses four color cameras to track the position of the ball. To find the position of the ball on an 
image, this system uses a reference orange color and convolves the resulting image with a circular
pattern using the fast Fourier Transform for efficiency. Instead of using the four cameras to output
one single 3D ball position, this system uses two pairs of two cameras. As a result, if all the cameras
are seeing the ball, two 3D position are estimated. In this system, it is not clear how to use more
cameras or how to determine which observations are reliable or not. Each table tennis
policy that used RTBlob had to implement its own outlier rejection heuristics to determine
which produced ball observations were reliable.

There are several other vision systems for robot table tennis, but none of them addresses the
problem of how to deal with mistakes from the object detection algorithm in the images.
Quick MAG 3~\cite{mag3} uses a motion blur and a ball trajectory model
to estimate and predict ball trajectories. In \cite{li2012ping}, a background model is used to 
extract the position of the ball. The detected blobs are filtered out according to their area, 
circularity and other factors. Finally, a ball model is used to predict the ball trajectory.
In~\cite{chen2015robust}, the authors focus on the physical models useful to predict the ball 
trajectory, and use these models for humanoid robot table tennis.

A common design pattern for all the discussed table tennis vision systems, is that the object
detection part consists on multiple heuristics based on background extraction, color finding
and basic shape matching on blobs. Although these approaches tend to work well in practice,
it is hard to adapt them to track different objects. Instead, we use machine learning 
methods for the object detection procedure. To track different objects, we only require 
to label new images by placing bounding boxes around the objects of interest and subsequently
retrain the system. 

\section{Reliable Real-Time Ball Tracking}
\label{sec:methods}

\begin{figure}
  \centering
  \includegraphics[height=2.5cm]{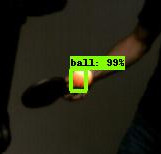}
  \includegraphics[height=2.5cm]{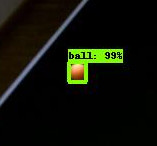}
  \includegraphics[height=2.5cm]{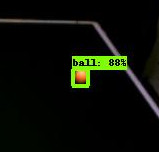}
  \caption{
    Ball detection with a Mobilnet deep network architecture using the Single Shot 
    Detection (SSD) method. 
    Note that the SSD method finds the location of the ball in all the images with relatively
    good accuracy. However, we obtain an average of 15 ball observations per second on a four
    camera setup, not efficient enough for a highly dynamic task like robot table tennis.
  }
  \label{fig:obj_det:ssd}
\end{figure}

End-to-end systems are an appealing strategy for system design in machine learning research, because
it makes less assumptions about how the system works internally. For our table tennis
vision setup, an end-to-end system should receive the input images from all the cameras and output
the corresponding ball location in 3D cartesian coordinates. However, such an end-to-end solution would
have a number of disadvantages for our table tennis setup. For example, adding new cameras or moving
around the existing cameras would require to re-train the entire system from scratch.

We divide our vision system into two subsystems. The object detection subsystem that outputs the ball
positions in pixel space for each image, and the position estimation subsystem that outputs a single 3D position 
of the ball based on a camera calibration procedure. To add new cameras we only need to run the calibration
procedure, and moving existing cameras requires only the re-calibration of the moved cameras.

First, we discuss about different methods used to detect objects in images.
In particular, we discuss about object detection and semantic segmentation methods. We show that although
both methods can successfully find table tennis balls in an image, the semantic segmentation method can
be used with smaller models, achieving the required real time execution requirements we need for
robot table tennis.

Subsequently, we discuss how to estimate a single 3D ball position from multiple camera observations. We
focus particularly on how to deal with erroneus estimates of the ball position in pixel space, for example,
when the object detection method fails and reports the location of some other object. We analyze the
algorithmic complexity of the proposed methods and we also provide execution times in a particular
computer for setups with different number of cameras.

\subsection{Finding the Position of the Ball in an Image}
\label{sec:methods:obj_det}

The problem of detecting the location of desired objects in images has been well studied in the 
computer vision community~\cite{huang2017speed}. Finding bounding boxes for objects in images is known
as object detection. In~\cite{liu2016ssd}, a method called Single Shot Detection (SSD) was proposed 
to turn a convolutional neural network for image classification into an object detection network.
An important design goal of the SSD method is computational efficiency. In combination with
a relatively small deep network architecture like Mobilnet~\cite{mobilenets}, designed for mobile
devices, it can perform real time object detection for some applications.

Figure~\ref{fig:obj_det:ssd}, shows example predictions of a Mobilnet architecture trained
with the SSD method in a ball detection data set. 
Each picture shows a section of the image with the corresponding bounding box
prediction. The resulting average processing speed using a GPU NVidia GTX 1080 was 60.2
frames per second on 200 x 200 pixel resolution images. For a 4 camera robot table 
tennis setup, this would result in about
15 ball observations per second. Unfortunately, for a high speed game like table tennis,
a significantly higher number of ball observations is necessary. However, we consider
important to mention the results we obtained with fast deep learning object detection 
techniques like the
SSD method, because it can be used with our method for a different application
where the objects to track are more complex and the required processing speeds are 
lower.

\begin{algorithm}[t]
  \begin{algorithmic}[1]
    \Require A probability image~$\matr{B}$, and a high and low
    thresholds~$T_h$ and~$T_l$.
    \Ensure A set of object pixels~$O$
    \State $(a,b) = \argmax_{(a,b)}{B_{ab}}$ \label{alg:find_ball_pixels:max_pix}
    \If { $B_{ab} < T_h$ }
      \State \Return~$\emptyset$
    \EndIf
    \State $O \gets \{(a,b)\}$ \label{alg:find_ball_pixels:flood:begin}
    \State $q \gets \queue(\{(a,b)\})$
    \While {$q$ is not empty}
      \State $\vect{x} \gets \pop(q)$
      \For { each neighbors $\vect{y}$ of $\vect{x}$ }
        \If { not $\vect{y} \in O$ and $B_{\vect{y}} > T_l$ }
          \State $\push(q,\vect{y})$
          \State $O \gets O \cup \{\vect{y}\}$
        \EndIf
      \EndFor
    \EndWhile \label{alg:find_ball_pixels:flood:end}
    \State \Return~$O$
  \end{algorithmic}
  \caption{Finding the set of pixels of an object.}
  \label{alg:find_ball_pixels}
\end{algorithm}

An alternative approach to find objects in images is to use a semantic segmentation 
method, where the output of the network is a pixelwise classification of the objects
of interest or background. For example, \cite{pohlen2017full}~uses deep 
convolutional neural networks to classify every pixel in a street scene
as one of 20 categories like car, person and road.
For our table tennis setup, a very simple and small model can be used considering 
that the ball has a simple spherical shape, small size and a relatively uniform 
color. To track the ball we only need two categories: Ball and Background.
We consider background anything that is not a table tennis ball. Let us denote
the resulting probability image as a matrix~$\matr{B}$, where~$B_{ij}$ is a scalar
denoting the probability that the pixel~$(i,j)$ of the original image corresponds
to a ball pixel or not.

In order to find the actual set of pixels corresponding to the ball, we need some kind 
of threshold based algorithm that makes a hard zero/one
decision of which pixels belong to the object of interest based on the obtained
probabilities. We used a simple algorithm that consists of finding the pixel
position~$(a,b)$ with maximum probability and a region of neighboring pixels
with a probability higher than a given threshold.

Algorithm~\ref{alg:find_ball_pixels} shows the procedure to obtain the set
of pixels corresponding to the ball from the probability image~$\matr{B}$.
The procedure receives two threshold values~$T_h$ and~$T_l$, that we call high
and low threshold respectively. In 
Line~\ref{alg:find_ball_pixels:max_pix}, we find the pixel position~$(a,b)$
with maximum probability on the probability image~$\matr{B}$. If the maximum
probability is lower than the high threshold value~$T_h$ we consider there
is no ball in the image and return an empty set of pixels. Otherwise, 
Lines~\ref{alg:find_ball_pixels:flood:begin} to~\ref{alg:find_ball_pixels:flood:end} 
find a region of neighboring pixels~$O$ around the maximum~$(a,b)$ with a 
probability larger than the low threshold~$T_l$ using a Breadth First Search 
algorithm.

The computational complexity of Algorithm~\ref{alg:find_ball_pixels} is linear on the number of 
pixels. If~$N_t$ represents the total number of pixels in the image and~$N_o$ the
number of pixels of the object to track, the computational complexity of 
Line~\ref{alg:find_ball_pixels:max_pix} alone in~$O(N_t)$ and the complexity
of the rest of the algorithm is~$O(N_o)$. However, Line~\ref{alg:find_ball_pixels:max_pix}
can be efficiently implemented in a GPU, whereas the rest of the algorithm is
harder to implement on a GPU due to its sequential nature. 
Given that~$N_t \gg N_o$, we decided to use the GPU to
execute Line~\ref{alg:find_ball_pixels:max_pix} and implemented the rest of the
algorithm in the CPU. In combination with the semantic segmentation approach
using a single convolutional unit, we obtained a throughput about 50 times
larger than the SSD method for our ball tracking problem.


\begin{figure}
  \centering
  \includegraphics[trim={10cm 9cm 6cm 2cm},clip,width=2.6cm]{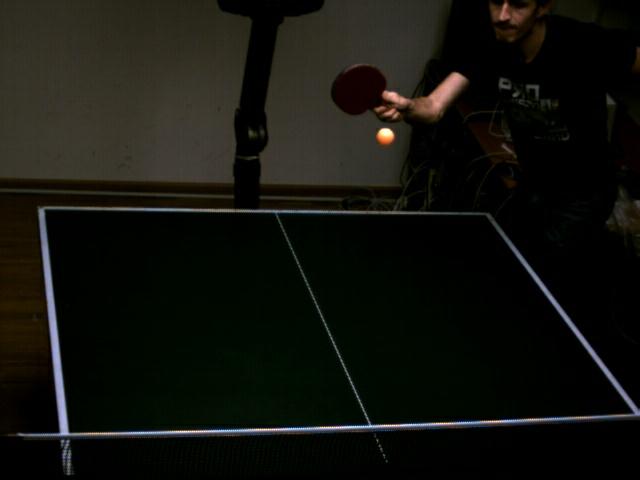}
  \includegraphics[trim={10cm 9cm 6cm 2cm},clip,width=2.6cm]{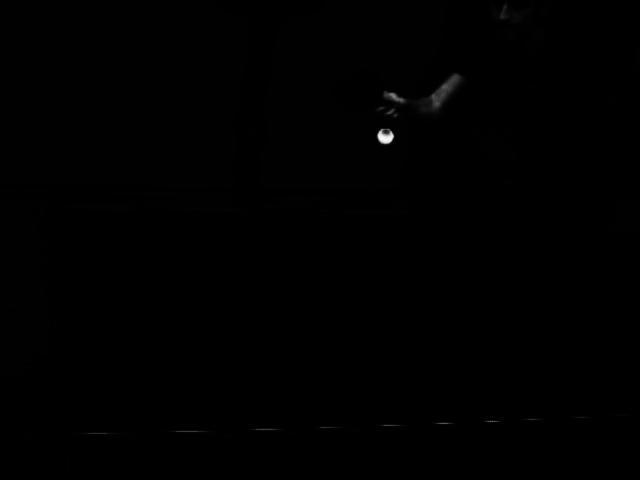}
  \includegraphics[trim={10cm 9cm 6cm 2cm},clip,width=2.6cm]{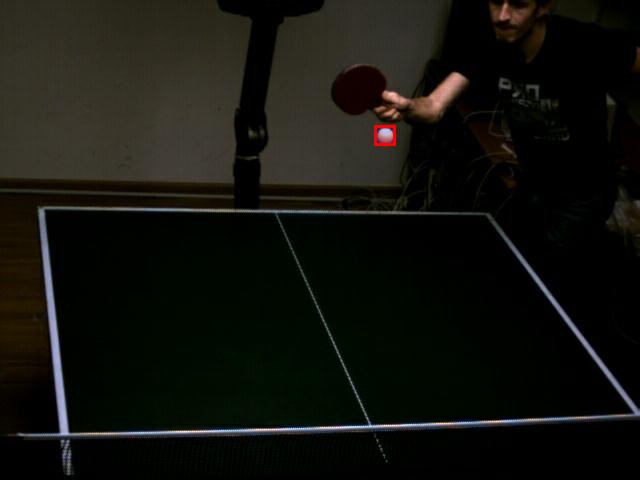}
  \caption{
    Ball detection using a single convolutional unit in a semantic segmentation 
    setting. The image on the left shows a section of a table tennis scene. 
    The image on the center shows the probability image~$\matr{B}$ representing
    the probability assigned to each pixel of being the ball. Dark means low
    probability and bright means high probability. 
    The image on the right shows the detected ball position. This
    simple model can successfully find the ball in the image, and it is around
    50 times faster than the SSD method.
  }
  \label{fig:obj_det:sseg}
\end{figure}

Figure~\ref{fig:obj_det:sseg} shows the semantic segmentation results for
the table tennis problem using a single convolutional unit with a 5x5
pixels filter size. The picture on the left shows a section of the
image captured with our cameras. The picture on the center shows
the probability image~$\matr{B}$ assigned by the model to each pixel as being the ball,
where white means high probability and black low probability. 
The picture on the right shows a bounding box that contains all pixels in~$O$
returned by Algorithm~\ref{alg:find_ball_pixels}.
Note that all the objects in the scene that are not the ball are
assigned by the model a very low probability of being the ball,
and most of the pixels of the ball are assigned a high
probability of being the ball. Actually, the only object that
can still be seen not completely dark in the probability image
is the human arm, because it has a similar color to the ball 
in comparison with the rest of the scene. 

The throughput of the single 5x5 convolutional unit is about 50
times higher that the throughput of the SSD method on the same
hardware with our implementations. As a result, we decided
to use the single convolutional unit as the ball detection method,
achieving the necessary ball observation frequency and accuracy
for robot table tennis. In Section~\ref{sec:experiments}, we 
analyze in detail the performance and accuracy of the single 
convolutional unit. In addition, we compare the accuracy
of our entire proposed system with the RTBlob vision system~\cite{rtblob}
and evaluate the playing performance of an existing robot table tennis 
method~\cite{gomez2018adaptation} using the proposed system.

\subsection{Robust Estimation of the Ball Position}
\label{sec:methods:pos_est}


Once we have the position of the ball in pixel space in multiple calibrated cameras, we
proceed to estimate a single reliable 3D ball position. The process to obtain an estimation
of the 3D position of an object given its pixel space position in two or more cameras is
called stereo vision. For an overview in stereo vision refer to~\cite{heyden2005multiple}.

\begin{algorithm}[t]
  \begin{algorithmic}[1]
    \Require A set of 2D observations and camera matrix 
      pairs~$S = \{\{x_{1},P{1}\},\dots,\{x_{k},P_{k}\}\}$, 
      and pixel error threshold~$\epsilon$.
    \Ensure A subset~$\hat{S} \subset S$ of maximal size without
      outliers.
    \State $\hat{S} \gets \emptyset$
    \For{ $i \in \{1,\dots,k-1\}$ }
      \For{ $j \in \{i+1,\dots,k\}$ }
        \State $\variable{candidate} \gets \stereo(\{P_i,P_j\}, \{x_i, x_j\})$
        \State $S_{ij} \gets \emptyset$
        \For { $k \in \{0,\dots,k\}$ }
          \State $\hat{x}_k \gets \project(\variable{candidate}, P_{k})$
          \State $\variable{p\_err} \gets \|x_{k} - \hat{x}_{k}\|_2$
          \If { $\variable{p\_err} < \epsilon$ }
            \State $S_{ij} \gets S_{ij} \cup \{x_k,P_k\}$ 
          \EndIf
        \EndFor
        \If { $|S_{ij}| > |\hat{S}|$ }
          \State $\hat{S} \gets S_{ij}$
        \EndIf
      \EndFor
    \EndFor
    \State \Return~$\hat{S}$
  \end{algorithmic}
  \caption{Remove outliers by finding the largest consistent subset of 
  2D observations for stereo vision.}
  \label{alg:max_inlier_set}
\end{algorithm}

We assume we have access to two functions~$\project$ and~$\stereo$
available from an stereo vision library.
Given a 3D point $X$ and a projection matrix~$P_i$ for the camera~$i$, the 
function~$x_i = \project(X,P_i)$ returns the pixel space coordinates~$x_i$ of projection 
of~$X$ in the image plane of camera~$i$. For the stereo vision method, we are given a set of
pixel space points~$\{x_1,\dots,x_k\}$ from~$k$ different cameras and their corresponding
projection matrices~$\{P_1,\dots,P_k\}$, and obtain an estimate of the 3D point~$X$
by
\[X = \stereo(\{x_1,\dots,x_k\},\{P_1,\dots,P_k\}).\] 
Intuitively, the
function~$\stereo$ finds the point~$X$ that minimize the pixel re-projection
error given by
\[ L(X) = \sum_{k}{\dist(x_k, \project(X,P_k))}, \]
where~$\dist$ is some distance metric like euclidean distance. 
If we could assume that the pixel space position of the balls is affected by 
independent Gaussian noise, taking all the available observations to minimize~$L(X)$ 
would yield the optimal solution. However, independent Gaussian noise is not
a valid assumption is the presence of outliers.

The algorithms described in Section~\ref{sec:methods:obj_det} to find the position
of the ball in the image will some times commit errors, reporting for example the 
position of other image objects as the ball. 
Assume that from a set~$S$ of pixel space ball observations reported 
by the vision system, some of the observations~$\hat{S} \in S$ are correctly
reported ball positions and the rest of the reported
observations~$\bar{S} = S - \hat{S}$ are erroneously reported ball positions.
We call~$\hat{S}$ the inlier set and~$\bar{S}$ the outlier set. We would like
to find the 3D ball position~$X$ that minimizes~$L(X)$ using only the set
of inliers~$\hat{S}$. Unfortunately, we do not know which observations from
the set~$S$ are outliers and which are inliers.

We define a set of pixel space observations as consistent if there is a 3D
point~$X$ such that~$L(X) < \epsilon$, where~$\epsilon$ is a pixel space 
error tolerance. We estimate~$\hat{S}$ by computing the largest subset of~$S$
that is consistent. The underlying assumption is that it should be hard to find
a single 3D position that explains a set of pixel observations containing 
outliers.
On the other hand, if the set of observations contains only inliers, we know it
should be possible to find a single 3D position~$X$, the cartesian position of the ball,
that explains all the pixel space observations. 

Algorithm~\ref{alg:max_inlier_set} shows the procedure we use to obtain the
largest consistent set of observations. Note that we need at least two cameras
to estimate a 3D position. Our procedure consists in trying all pairs of 
cameras~$(i,j)$, estimating a candidate 3D position only with those two 
observations, and subsequently counting how many cameras are consistent
with the estimated candidate position. If~$c$ represents the number of cameras
reporting a ball observation, the computational complexity of this algorithm 
is~$O(c^3)$. 

For a vision system of less than 30 cameras, we obtained real time performance
even using a sequential implementation of Algorithm~\ref{alg:max_inlier_set}.
Nevertheless, it is easy to parallelize Algorithm~\ref{alg:max_inlier_set}. 
Note that the outermost two for loops can be run independently in
parallel. In Section~\ref{sec:experiments}, we evaluate the real time performance and the
accuracy of the 3D estimation simulating scenarios with different number of 
cameras and probability of outliers. Afterwards, we evaluate the error in
the real system and compare it with the RTBlob method on the same experimental setup.

\section{Experiments and Results}
\label{sec:experiments}

We evaluate the proposed system in a simulation environment and in a real robot platform.
In simulation, we measure the accuracy of the system as we increase the number
of cameras and when we change the probability of obtaining outliers.
We use the real robot platform to evaluate the
interaction of all the components of the proposed system. In particular, we measure
the accuracy and robustness of the proposed system and compare it with the RTBlob
method. In addition, we evaluate the success rate of a method proposed 
in~\cite{gomez2018adaptation} to return balls to the opponent's court with the proposed
vision system. We have a slightly higher success rate using the proposed vision system
than using the RTBlob system even after removing all the outlier rejection heuristics
implemented in~\cite{gomez2018adaptation}.

\subsection{Evaluation on a Simulation Environment}

\begin{table}
  \centering
  \begin{tabular}{@{}c|cccccc@{}}
    \toprule
    \multirow{2}{*}{$c$} & & \multicolumn{5}{c}{\bf{Probability of Outliers~$p_o$}} \\
     & & 1\% & 5\% & 10\% & 25\% & 50\% \\
    \midrule
    \multirow{2}{*}{4} & E & 0.71 cm & 0.85 cm & 0.84 cm & 0.79 cm & 4.67 cm \\
    & F & 0.1\% & 0.5\% & 2.0\% & 9.7\% & 37.7\% \\
    \midrule
    \multirow{2}{*}{8} & E & 0.52 cm & 0.53 cm & 0.59 cm & 0.94 cm & 6.84 cm \\
    & F & 0.0\% & 0.0\% & 0.0\% & 0.1\% & 4.5\% \\
    \midrule
    \multirow{2}{*}{15} & E & 0.35cm & 0.36 cm & 0.37 cm & 0.41 cm & 4.72 cm \\
    & F & 0.0\% & 0.0\% & 0.0\% & 0.0\% & 0.02\% \\
    \midrule
    \multirow{2}{*}{30} & E & 0.24cm & 0.25 cm & 0.25 cm & 0.28 cm & 0.35 cm \\
    & F & 0.0\% & 0.0\% & 0.0\% & 0.0\% & 0.0\% \\
    \bottomrule
  \end{tabular}
  \caption{
    Estimation error (E) and failure probability (F) of the 3D position estimation
    procedure in the presence of outliers. A failure means that the system does not
    report any ball position at all because the maximum consistent set returned by 
    Algorithm~\ref{alg:max_inlier_set} consisted of less than two ball observations.
    Otherwise, the system return an estimated ball position and we report the distance
    in centimeters to the ground truth position. We simulate multiple scenarios with a different
    number of cameras and different probability of outliers. Note that as the number of
    cameras increases and the probability of obtaining outliers decreases the system becomes 
    more reliable.
  }
  \label{tab:sim_outliers}
\end{table}

To evaluate the proposed methods in scenarios that include different number of cameras and probability
of outliers, we use a simulation scenario. The advantage of evaluating in simulation is that we have
access to exact ground truth data and we can easily test the robustness and accuracy of the system.
In this section, we evaluate the robustness of the introduced procedure to find the 3D position of the ball from
several unreliable pixel space observations. First, we want to evaluate the performance of 
Algorithm~\ref{alg:max_inlier_set} independently of the rest of the system. In addition, we want to test the accuracy
and running time of the algorithm for different amount of cameras and outlier rates. 

The simulation for a scenario with $c$ cameras and a probability of outlier~$p_o$ consists of the following 
steps: First, we generate randomly a 3D ball position~$X$ in the work space of the robot and project it to each 
camera using the calibration matrices. We add a small Gaussian noise with a standard deviation of 1.3 pixels to
the projected pixel space position, because that is the average re-projection error reported by the camera
calibration procedure. For each camera, we replace the obtained pixel space position by some other random
position in the image plane with probability~$p_o$. Subsequently, we attempt to obtain the 3D ball position
with Algorithm~\ref{alg:max_inlier_set}. If it fails to obtain any position at all, we count it as a
failure. Otherwise, we measure the error of the obtained position with the ground truth value~$X$.

Table~\ref{tab:sim_outliers} shows the results for scenarios with a number of cameras
ranging from 4 to 30 and probability of outliers ranging from 1\% to 50\%. For every
combination of number of cameras and probability of outliers, we report the 
failure rate~(F) and the error~(E) between the ground truth position and the reported
ball position. As the probability of outliers increases the error and failure rate
increases as it is expected. Similarly, as more cameras are added to the system,
the robustness of the system increases, obtaining smaller errors and failure
rates. There are few entries in Table~\ref{tab:sim_outliers} that seem to contradict
the trend to reduce the error as more cameras are introduced or the outlier rate
drops. For example, for an outlier rate of 50\% the error with four cameras is 4.67 cm
whereas the error for eight cameras is 6.84 cm. Note however that the failure rate
for four cameras is much higher than for eight cameras in this case. 

\begin{table}
  \centering
  \begin{tabular}{@{}lcccccc@{}}
    \toprule
    \bf{Cameras} & 4 & 8 & 15 & 30 & 50 \\
    \bf{Run time (ms)} & 0.001 & 0.012 & 0.015 & 3.02 & 11.46 \\
    \bottomrule
  \end{tabular}
  \caption{
    Run time in milliseconds of a sequential implementation of 
    Algorithm~\ref{alg:max_inlier_set} with respect to the number of cameras.
    For a system of up to 30 cameras, the sequential implementation of
    Algorithm~\ref{alg:max_inlier_set} provides real time performance for
    more than 200 ball observations per second.
  }
  \label{tab:outlier:runtime}
\end{table}

Adding more cameras to the system improves accuracy and robustness. However, it
also increases the computation cost. The cost of the image processing part grows
linearly with the number of cameras, but can be run independently in parallel for
every camera if necessary. Therefore, we will focus on the cost of the position
estimation procedure as the number of cameras grows. As discussed in 
Section~\ref{sec:methods:pos_est}, the cost of the position estimation procedure 
in~$O(c^3)$. Table~\ref{tab:outlier:runtime} shows the run time in milliseconds of
a sequential implementation of~\ref{alg:max_inlier_set} in C++ in a Lenovo Thinkpad
X2 laptop. For a target frequency rate of 200 observations per second we need a processing
time smaller than 5 milliseconds. Note that even the sequential implementation of
Algorithm~\ref{alg:max_inlier_set} has the required real time performance for systems
up to 30 cameras. In addition, Algorithm~\ref{alg:max_inlier_set} can be easily
parallelized if necessary as discussed in Section~\ref{sec:methods:pos_est}.

It is important to note that on a real system not all the cameras might be seeing
the work space of the robot. For example, in the real robot setup we used four cameras
but there are many parts of the work space that are covered only by two cameras,
reducing the effective robustness of the system on those areas. However, the outlier
rate of the image processing algorithms is below 1\% in practice, and good results can 
be obtained using a small number of cameras as we discuss in the next section.


\subsection{Evaluation on the Real Robot Platform}

We evaluate the entire proposed system in the real robot platform and compare
the performance to the RTBlob system presented in~\cite{rtblob}. The
evaluation on the real robot platform consisted of two experiments. First,
we attach a table tennis ball to the robot end effector. We move the
robot and use its kinematics to compute the position of the ball and
use it as ground truth to compare against the ball positions obtained
with the vision system. Finally, we evaluate the playing performance of
the robot table tennis strategy introduced in~\cite{gomez2018adaptation}
if we remove all the heuristics used to remove vision outliers.

We compare the performance of the proposed vision system with RTBlob~\cite{rtblob}.
The RTBlob system has been used for robot table tennis 
experimentation~\cite{gomez2016using,huang2015learning}. In order to compare
the accuracy of both systems, we need access to ground truth positions. We use
the joint sensors of the robot and the robot kinematics to compute the Cartesian
position of the robot end effector. We attach the ball to the robot end effector
and use the Cartesian position computed with the joint measurements as ground truth.

\begin{figure}
  \centering
  \setlength\figurewidth{8.2cm}
  \setlength\figureheight{6cm}
  \setlength\axislabelsep{0mm}
  \input{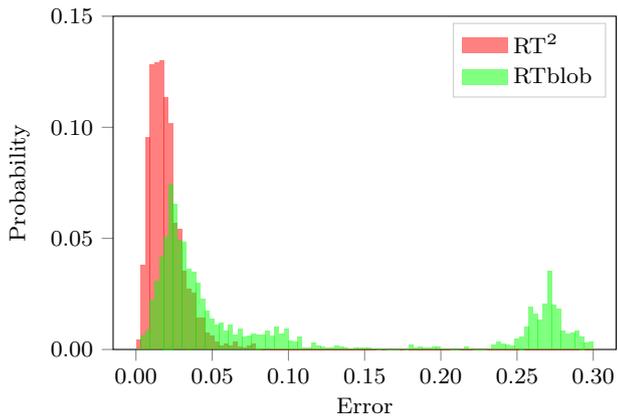}
  \caption{
    Histogram of the error of the presented vision system and RTBlob. We call the
    proposed vision system \sysname{}, depicted in red in the histogram. The ball
    is attached to the robot end effector and the end effector position computed with
    the kinematics is used as ground truth. The accuracy of the proposed vision system
    is superior to the RTBlob system. In terms of robustness to outliers, the proposed
    system (\sysname{}) outperforms the RTBlob system as expected. The error distribution for our
    system is unimodal, whereas the RTBlob system error is multimodal, reflecting
    the sensitivity of the RTBlob system to the presence of outliers.
  }
  \label{fig:vision_comp}
\end{figure}

Figure~\ref{fig:vision_comp} shows a histogram of the error of the RTBlob method and
the method proposed in this paper. We called the proposed system \sysname{} in the figure, 
standing for Real Time Reliable Tracking.
The error is computed as the distance between the position
reported by the vision and the ground truth computed with the robot kinematics. Note
that the proposed vision system outperforms the RTBlob method in terms of accuracy, but
specially in terms of outliers. The distribution of errors for \sysname{} concentrates
the probability mass between 0 cm and 5 cm error. On the other hand, the error distribution
of the RTBlob method is multimodal. The first mode corresponds to the scenario where
all the cameras detected the ball correctly, and in this case the error mass is also concentrated
below a 7 cm error threshold. The second mode shows a high probability of error between 25 cm and
30 cm, and it is likely to correspond to a scenario where one of the four cameras reported an incorrect
ball position.

During the execution of the accuracy experiment reported in Figure~\ref{fig:vision_comp}, the
system proposed in this paper never reported any ball position whose error was larger than 10 cm.
On the other hand, the RTBlob system reported errors on the order of tens of meters
with probability around 0.1\%. As a result, the table tennis strategies that use the RTBlob
method have to incorporate strategies to filter outliers to work properly.

In the last part of this section, we present a final experiment where we use the proposed
vision system to return table tennis balls with the robot to the opponent's court. We use
a method presented in~\cite{gomez2018adaptation}, that is based on Probabilistic Movement 
Primitives (ProMPs) and learning from a human teacher. The system presented 
in~\cite{gomez2018adaptation} was originally designed to use the RTBlob method as the vision
system. To detect and filter outliers, the RANSAC algorithm was used on a set of
initial observations fitting a second order polynomial. Once a set of candidate positions
is found, a Kalman filter is used to predict the ball trajectory and subsequent ball
observations are rejected if they are more than 3 standard deviations away from the
mean position predicted by the Kalman filter.

We decided to remove the heuristics to filter outliers, accepting all ball observations
as valid, and test the method with the proposed vision system. We define "success" as
the robot hitting the incoming ball and sending it back to the opponent's court according
to the table tennis rules. The average success rate using the RTBlob vision system and
all the outlier rejection heuristics was of {\bf 68 \%}, whereas using the proposed
vision system and no outlier rejection heuristics the average success rate was {\bf 70 \%}.
Given the variability of the table tennis performance between multiple experiments, we can
not say that the improvement with the new vision system is statistically significant.
However, we find remarkable that the success rate of table tennis strategy presented
in~\cite{gomez2018adaptation} did not decrese after the outlier rejection
heuristics were removed. We think that the slight improvement on the success rate by
using the proposed vision system is due to the improved frame rate, that is about
3 times as high as that of the RTBlob implementation provided by the authors~\cite{rtblob}.

\section{Conclusions and Discussion}
\label{sec:conclusion}

This paper introduces a vision system for robot table tennis focused on reliability
and real time performance. The implemented system is released as an open source 
project~\cite{ball_track_impl} to facilitate its usage by the community.
The proposed vision system can be easily adapted for different
tracking tasks by labeling a new data set and training the object detection algorithm.
For the object detection part, this paper evaluates two different approaches used commonly 
in the computer vision community that are known for obtaining real time performance. We decided
to use the simpler approach for tracking table tennis balls due to its high throughput.

For the position estimation procedure we proposed an algorithm that focuses on reliability,
by assuming that some times the object detection methods will report wrong ball positions
on the provided images. We evaluate the proposed method thoroughly in simulation and in
the real robot platform. In simulation, we test the accuracy of the system under different
probability of outliers and number of cameras. In the real system, we evaluate the complete
proposed system in a four camera setup and compare it with the RTBlob vision system. We
show that our system provides higher accuracy, and outperforms the RTBlob system in
robustness to outliers. Finally, we test an existing technique to return table tennis
balls to the opponent's court with our vision system. We removed all the outlier detection
techniques used by the table tennis algorithm and obtained a small increase in success rate
compared to the RTBlob system with all the outlier detection techniques present. We believe 
the proposed approach will help future research in robot table tennis by allowing the
researchers to focus on the table tennis policies instead of techniques to deal with an 
unreliable vision system.


\bibliographystyle{plain}
\bibliography{refs}   

\end{document}